\documentclass[10pt,twocolumn,letterpaper]{article}

\usepackage{cvpr}
\usepackage{times}
\usepackage{epsfig}
\usepackage{graphicx}
\usepackage{amsmath}
\usepackage{amssymb}
\usepackage{caption}
\usepackage{subcaption}
\usepackage{booktabs}
\usepackage{multirow}
\usepackage{graphics}
\usepackage{bigstrut}
\usepackage{enumitem}

\usepackage[pagebackref=true,breaklinks=true,letterpaper=true,colorlinks,bookmarks=false]{hyperref}

 \cvprfinalcopy 

\def\cvprPaperID{1737} 

\newcommand{\Section}[1]{\vspace{-4pt}\section{#1}\vspace{-4pt}}
\newcommand{\Subsection}[1]{\vspace{-3pt}\subsection{#1}\vspace{-3pt}}

\newenvironment{tight_itemize}{\begin{itemize}[leftmargin=*] \itemsep
-3pt}{\end{itemize}}

\ifcvprfinal\fi
\begin{document}

\title{ASP Vision: Optically Computing the First Layer of Convolutional Neural Networks using Angle Sensitive Pixels}

%

\author{Huaijin G. Chen\footnotemark[1]\\
Rice University\\
\and
Suren Jayasuriya\footnotemark[1]\\
Cornell University\\
\and
Jiyue Yang\\
Cornell University\\
\and
Judy Stephen\\
Cornell University\\
\and
Sriram Sivaramakrishnan\\
Cornell University\\
\and
Ashok Veeraraghavan\\
Rice University\\
\and
Alyosha Molnar\\
Cornell University\\
}

\maketitle
\footnotetext[1]{Authors contributed equally to this paper}

\begin{abstract}
	Deep learning using convolutional neural networks (CNNs) is quickly becoming the state-of-the-art for challenging computer vision applications. However, deep learning's power consumption and bandwidth requirements currently limit its application in embedded and mobile systems with tight energy budgets. In this paper, we explore the energy savings of optically computing the first layer of CNNs. To do so, we utilize bio-inspired Angle Sensitive Pixels (ASPs), custom CMOS diffractive image sensors which act similar to Gabor filter banks in the V1 layer of the human visual cortex. ASPs replace both image sensing and the first layer of a conventional CNN by directly performing optical edge filtering, saving sensing energy, data bandwidth, and CNN FLOPS to compute. Our experimental results (both on synthetic data and a hardware prototype) for a variety of vision tasks such as digit recognition, object recognition, and face identification demonstrate a reduction in image sensor power consumption and data bandwidth from sensor to CPU, while achieving similar performance compared to traditional deep learning pipelines.

\end{abstract}



\begin{figure}[t]
\begin{center}
   \includegraphics[width=.5\textwidth]{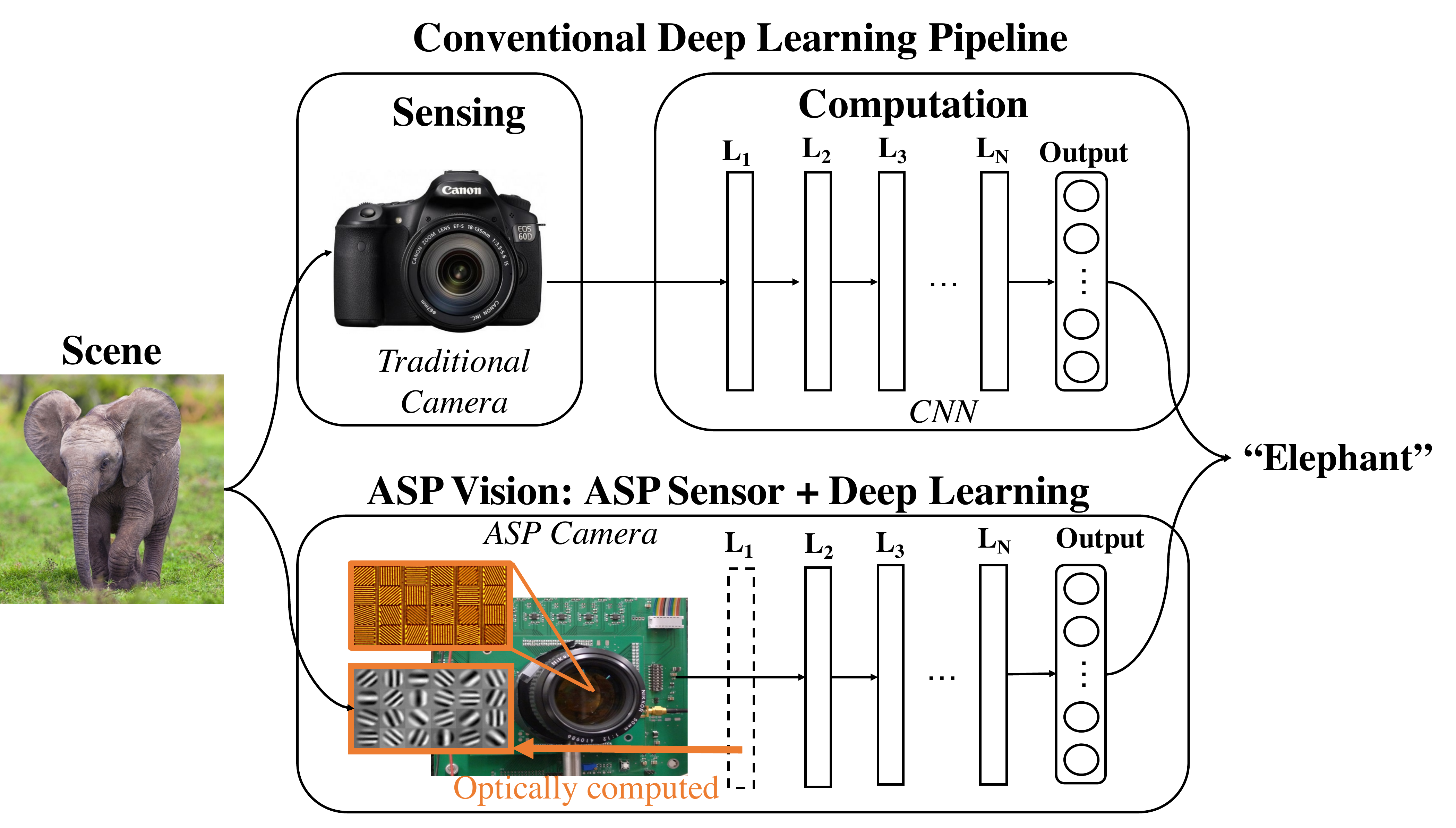}
\end{center}
   \caption{\textbf{A diagram of our proposed ASP Vision}. ASP Vision, our proposed system, is compared with a conventional deep learning pipeline. ASP Vision system saves energy and transmission bandwidth in the sensing stage, compared to a traditional camera.}
\label{fig:systemdiagram}
\end{figure}

\Section{Introduction}
	State-of-the-art visual recognition algorithms utilize convolutional neural networks (CNNs) which use hierarchical layers of feature computation to discriminate visual stimuli. Early CNNs from LeCun et al.~\cite{lecun1998gradient} showed promising results in digit recognition~\cite{lecun1998mnist}. The advent of GPU computing has allowed CNN training on large, public online data sets, and triggered an explosion of current research. CNNs have started to perform on par with or even surpass humans on some image recognition challenges such as ImageNet~\cite{krizhevsky2012imagenet, russakovsky2014imagenet}. CNNs have been universally applied to different vision tasks including object detection and localization~\cite{szegedy2014going}, pedestrian detection~\cite{szarvas2005pedestrian}, face recognition~\cite{SchroffKP15}, and even synthesizing new objects~\cite{dosovitskiy2015learning}. However, many applications in embedded vision such as vision for mobile platforms, autonomous vehicles/robots, and wireless sensor networks have stringent constraints on power and bandwidth, limiting the deployment of CNNs in these contexts. 

\Subsection{Motivation and Challenges}
Porting deep learning vision systems to embedded and battery-operated applications necessitates overcoming the following challenges:

\begin{tight_itemize}
\item{\textbf{Sensor power:} Image sensors are notoriously power-hungry, sometimes accounting for more than 50\% of the power consumption in many embedded vision applications~\cite{likamwa2013energy}. In addition, current image sensors are not optimized to significantly save power for such computer vision tasks~\cite{likamwa2013energy}. Several researchers, most recently~\cite{likamwaRedEye}, have argued that always-on, battery-operated, embedded vision systems necessitate a complete redesign of the image sensor to maximize energy-efficiency. }

\item{\textbf{Computing power:} CNNs, while providing enormous performance benefits, also suffer from significantly increased computational complexity. GPUs and multi-core processors are power hungry, and the number of FLOPS (floating point operations) for CNNs can easily be on the order of billions.}

\item{\textbf{Data bandwidth:} Data bandwidth requirements place strict design constraints on traditional vision architectures. Moderate image resolution of $1$ megapixel at $30$ fps (frames per second) results in a bandwidth requirement of over $0.5$ Gbps (Giga-bits per second). This can bottleneck I/O buses that transfer images off the sensor to the CPU and increases the power requirements, computational complexity, and memory for the system.}
\end{tight_itemize}

\Subsection{Our Proposed Solution}
To solve the challenges described above, we explore novel image sensors that can save energy in an embedded vision pipeline. In particular, we use existing Angle Sensitive Pixels (ASPs)~\cite{wang2012light}, bio-inspired CMOS image sensors that have Gabor wavelet impulse responses similar to those in the human visual cortex, to perform optical convolution for the CNN first layer. We call this combination of ASP sensor with CNN backend \textit{ASP Vision}. This system addresses embedded deep learning challenges by: 
\begin{tight_itemize}
\item{\textbf{Reducing sensor power} by replacing traditional image sensors with energy-efficient ASPs that only digitize edges in natural scenes. }
\item{\textbf{Reducing computing power} by optically computing the first convolutional layer using ASPs, thus leaving subsequent network layers with reduced FLOPS to compute.}
\item{\textbf{Reducing bandwidth} by relying on the inherent reduced bandwidth of ASP sensors encoding only edge responses.}
\end{tight_itemize}

\Subsection{Contributions}
In this paper, we will describe in detail our system for optically computing the first layer of CNNs. Note that we are neither introducing ASPs for the first time nor claiming a new CNN architecture. Instead, we are deploying ASPs to increase energy efficiency in an embedded vision pipeline while maintaining high accuracy. In particular, our main contributions in this paper include:  
\begin{tight_itemize}
\item{Showing the optical response of Angle Sensitive Pixels emulates the first layer of CNNs }
\item{Analysis of the energy and bandwidth efficiency of this optical computation}
\item{Evaluation of system performance on multiple datasets: MNIST~\cite{lecun1998mnist}, CIFAR-10/100~\cite{krizhevsky2009learning}, and PF-83~\cite{becker2013evaluating}.}
\item{An operational prototype of the ASP Vision system and real experimental results on digit recognition and face recognition using our prototype.}
\end{tight_itemize}

\Subsection{Limitations}
Our proposed approach is also limited by some practical factors. While there are significant potential FLOPS savings from optical computation, our current prototype achieves a modest fraction of these savings due to the prefabricated sensor's design choices. In addition, ASPs themselves are challenged with low light efficiency and reduced resolution that we address in detail in Section 3.4.2. Finally, our current hardware prototype has limited fidelity since it was not fabricated in an industrial CMOS image sensor process. We discuss this in Section 5. We caution readers from placing too much expectation on the visual quality of a research prototype camera, but hope the ideas presented in the paper inspire further research in novel cameras for computer vision systems.


\Section{Related Work}
In this section, we survey the literature with particular focus on energy-efficient deep learning, computational cameras, and hardware-based embedded vision systems.

\textbf{Convolutional Neural Networks} are currently the subject of extensive research. A high level overview of CNNs is given by LeCun et al.~\cite{lecun2015deep}. Since this paper does not improve CNN accuracy or propose new networks, we highlight recent work on real-time performance and resource efficiency. Ren et al. use faster R-CNN~\cite{ren2015faster} to achieve millisecond execution time, enabling video frame rates for object detection. In addition, researchers have explored reducing floating point multiplications~\cite{LinCMB15}, quantization of weights in CNNs~\cite{gong2014compressing, HanMD15}, network compression~\cite{weinbergercompressed2015}, and trading off accuracy for FLOPs~\cite{SchroffKP15}. 

On the sensor side, \textbf{computational cameras} have emerged to expand the toolset of modern imaging systems. Cameras have been augmented to capture light fields~\cite{hirsch2014switchable, ng2005light,veeraraghavan2007dappled}, polarization~\cite{gruev2010ccd}, high dynamic range~\cite{serrano2016convolutional}, and depth~\cite{Payne2014}. Similar to ASPs, cameras that compute features include on-chip image filtering~\cite{gruev2002implementation, nilchi2009focal} or detect events~\cite{lichtsteiner2008128}.  

\textbf{Embedded vision} has been spurred by advances in imaging technology and digital processing. For convolutional neural networks, analog ASICs~\cite{boser1991analog}, FPGAs~\cite{farabet2009fpga}, and neuromorphic chips~\cite{pham2012neuflow} implement low power calculations with dedicated hardware accelerators. LiKamWa et al.~\cite{likamwaRedEye} propose a new analog-to-digital converter for image sensors that performs CNN image classification directly to avoid the I/O bottleneck of sending high resolution images to the processor. Micro-vision sensors~\cite{koppal2013toward} perform optical edge filtering for computer vision on tight energy budgets. Similar to this paper, inference/learning on coded sensor measurements from compressive sensing imaging has saved bandwidth/computation~\cite{iliadis2016deep, kulkarni2015reconstruction, lohit2015reconstruction}. Dynamic Vision Sensors (DVS) have been used for face recognition while saving energy compared to conventional image sensors~\cite{miyatani2016wacv}.  All this research forecasts higher levels of integration between deep learning and embedded vision in the future.
\vspace{-1mm}

\begin{figure}[t]
\begin{center}
   \includegraphics[width=.5\textwidth]{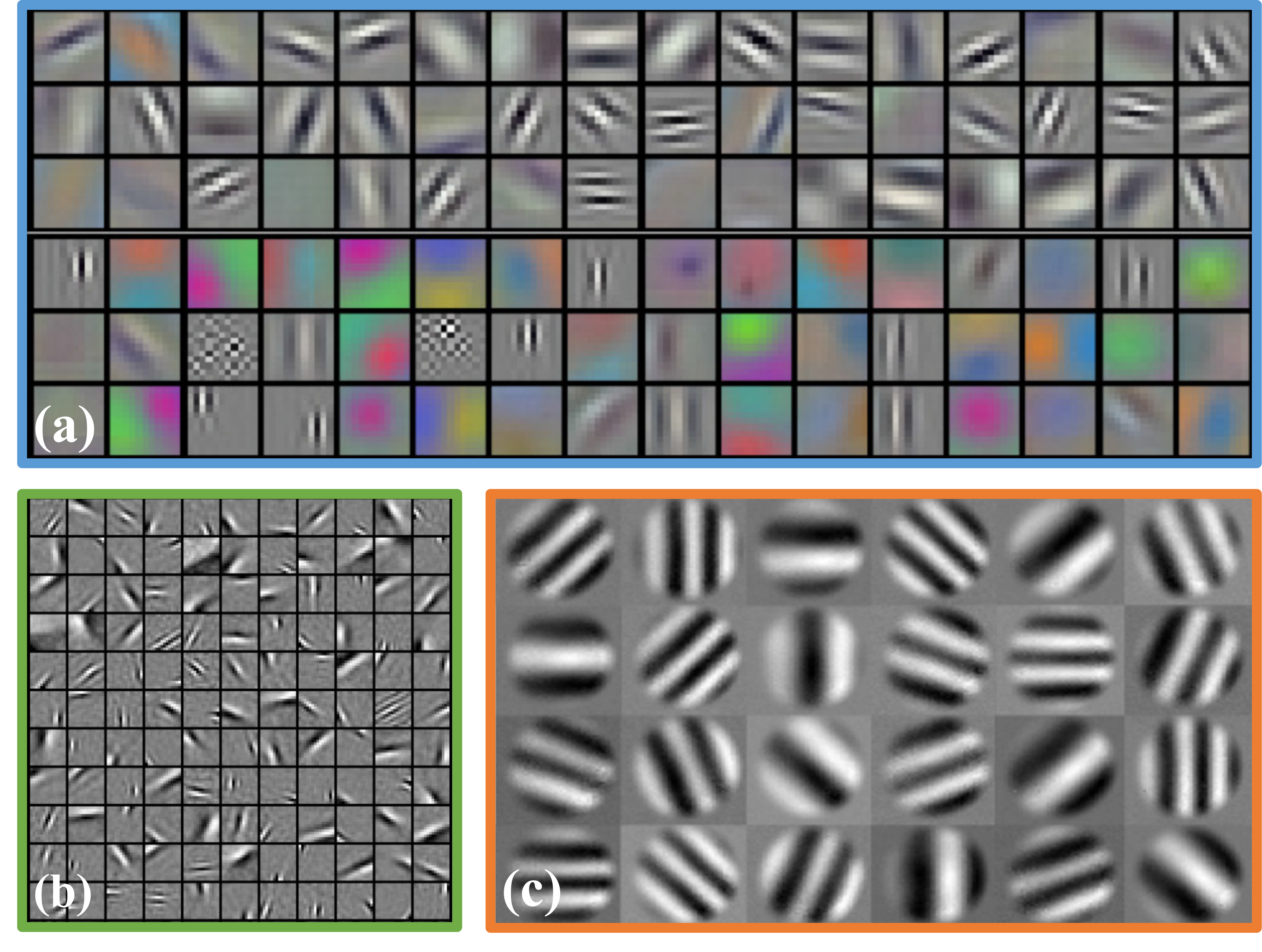}
\end{center}
   \caption{\textbf{Comparison of first layer weights for three different systems:} (a) Traditional deep learning architecture AlexNet trained on ImageNet~\cite{krizhevsky2012imagenet}, (b) Set of weights given by sparse coding constraints similar to the receptive fields of simple cells in the V1~\cite{olshausen1996emergence}, and (c) ASP optical impulse responses for 2D incidence angles~\cite{hirsch2014switchable}.}
\label{fig:V1}
\end{figure}

\Section{ASP Vision}
   \vspace{-0.5mm}
A diagram of our proposed ASP Vision system is presented in Figure \ref{fig:systemdiagram}. The custom image sensor is composed of Angle Sensitive Pixels which optically computes the first layer of the CNN used for visual recognition tasks. In the following subsections, we describe how hardcoding the first layer is application-independent, ASP design, and how ASPs perform optical convolution with energy and bandwidth savings. Finally, we discuss current limitations with ASP design and imaging for embedded vision.
\subsection{Hardcoding the First Layer of CNNs}
     In partitioning a deep learning pipeline, a central question is what layers of the CNN should be implemented in hardware versus software. Hardcoded layers generally lead to significant energy savings provided a suitably efficient hardware implementation is used. However, maintaining the last layers of CNNs in software allows flexibility for network reconfiguration, transfer learning~\cite{pan2010survey}, and fine-tuning~\cite{bengio2012deep}. 

In our system, we are interested in optically computing the first layer of CNNs in hardware. We note recent research that shows the first layers of CNNs are application-independent and transferable~\cite{yosinski2014transferable}. Fine-tuning the CNN by retraining only the last few layers on a new application-domain leads to high accuracy. In particular, the first layer learned by most CNN architectures consists of oriented edge filters, color blobs, and color edges (as visualized AlexNet's ~\cite{krizhevsky2012imagenet} first layer in Figure \ref{fig:V1}(a)). These edge filters are not a surprise and are also found in the receptive fields of simple cells in the V1 layer of the human visual system. Olhausen and Field characterized these filters as Gabor wavelets, visualized in Figure \ref{fig:V1}(b), and showed how they perform sparse coding on natural image statistics~\cite{olshausen1996emergence}. 

Therefore hardcoding this first layer should be independent of application and roughly converges to the same set of Gabor filters for most networks. Our main idea is to use Angle Sensitive Pixels (ASPs) in our image sensor front end to compute this convolutional layer in the optical domain at low electronic power consumption.

\subsection{Angle Sensitive Pixels}
	\subsubsection{Background}
ASPs are photodiodes, typically implemented in a CMOS fabrication process, with integrated diffraction gratings that image the Talbot diffraction pattern of incoming light~\cite{wang2012light}. These diffraction gratings give a sinusoidal response to incident angle of light given by the following equation~\cite{hirsch2014switchable}: \begin{equation}
	i(x,y) = 1+m\cos \left( \beta \left(\cos\left( \gamma \right) \theta_x + \sin\left(\gamma\right) \theta_y \right)  + \alpha \right),
	\label{eq:aspresponse_2D}
\end{equation}
 where $\theta_x, \theta_y$ are 2D incidence angles, $\alpha, \beta, \gamma$ are parameters of the ASP pixel corresponding to phase, angular frequency, and grating orientation, and $m$ is the amplitude of the response. A tile of ASPs contain a diversity of angle responses, and are repeated periodically over the entire image sensor to obtain multiple measurements of the local light field. ASPs have been shown to capture 4D light fields~\cite{hirsch2014switchable} and polarization information~\cite{jayasuriya2015dual}. An advantage of these sensors is that they are CMOS-compatible and thus can be manufactured in a low-cost industry fabrication process. 
\vspace{-1mm}

\subsubsection{Optical Convolution}
\begin{figure}[t]
\begin{center}
   \includegraphics[width=1\linewidth]{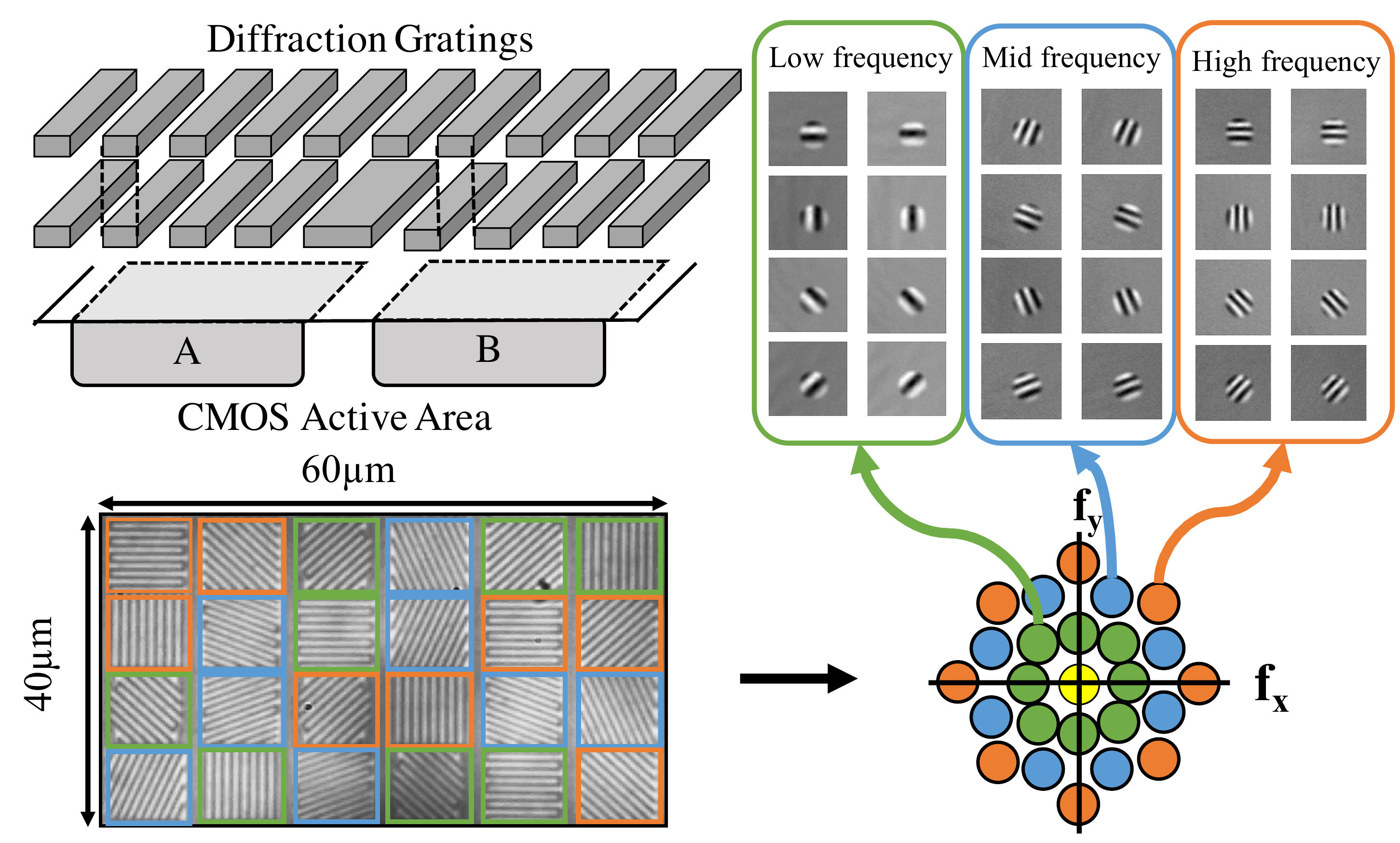}
\end{center}
   \caption{\textbf{ASP Pixel Designs:} ASP differential pixel design using diffraction gratings is shown. A 4 $\times$ 6 tile contains 10$\mu$m pixels whose optical responses are Gabor filters with different frequency, orientation, and phase. These filters act as bandpass filters in 2D frequency space~\cite{wang2012compression}.}
\label{fig:ring}
\end{figure}

\begin{figure}[t]
\begin{center}
   \includegraphics[width=.9\linewidth]{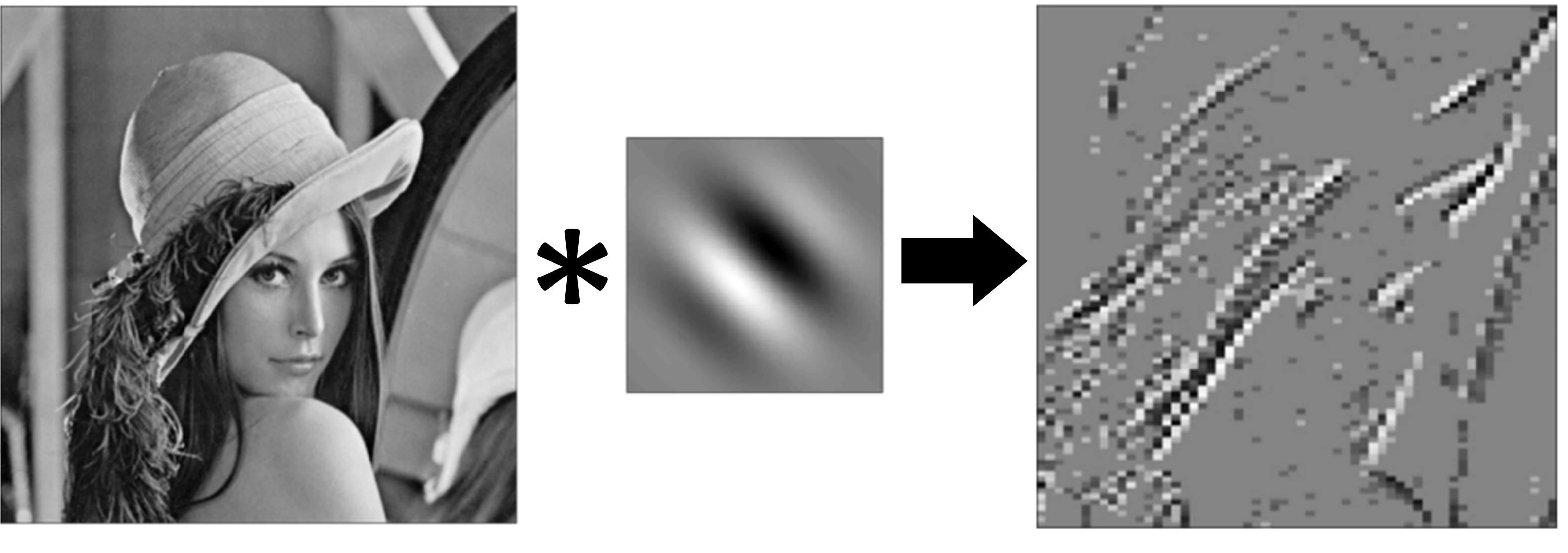}
\end{center}
   \caption{\textbf{ASP Differential Output:} Optical convolution of a scene with a differential ASP impulse response results in an edge filtered image (real images from prototype camera in~\cite{wang2012compression}).}
\label{fig:aspfilter}
\end{figure}

In particular, ASP responses to incidence angle allow optical convolution and edge filtering. Using two differential pixels of phase $\alpha$ and phase $\alpha+\pi$ (pixels A and B of Figure \ref{fig:ring}), we subtract their responses, $i_{\alpha} - i_{\alpha+\pi}$, to obtain the sinusoidal term of Equation 1 which depends solely on angle without the fixed DC offset. Figure \ref{fig:ring} shows these measured differential pixel's impulse responses across an ASP tile. They resemble several different Gabor wavelets of different frequency, orientation, and phase which tile 2D frequency space. These impulse responses are convolved optically with objects in the scene during the capture process. The resulting ASP output correspond to edge filtered images as displayed in Figure \ref{fig:aspfilter}. We use this optical convolution with Gabor wavelets to compute the first convolutional layer of a CNN.

Analogously, the V1 layer of the visual cortex contains Gabor wavelet responses for the receptive fields of simple cells, and Olhausen and Field showed that this representation is maximally efficient in terms of information transfer and sparse coding~\cite{olshausen1996emergence}. This is partly why we claim this system is bio-inspired: we are taking advantage of ASP's Gabor-like response to naturally compress the statistics of natural scenes to edges. This edge representation has direct implications for the low power consumption of ASPs.

\subsubsection{Energy and Bandwidth Efficiency}
Prior work has designed ASP readout circuitry to leverage the sparseness of edge filtered images, enhancing energy efficiency~\cite{wang2012compression}. Circuit readout that involves a differential amplifier can read out differential pixels, subtract their responses, and feed it to an analog-to-digital (ADC) converter~\cite{wang2012compression}. This ADC is optimized to only convert pixels when there is sufficient edge information, leading to low power image sensing and digitization as compared to a traditional image sensor. 

A comparison of an ASP-based image sensor~\cite{wang2012compression} to a modern Sony mobile image sensor~\cite{suzuki2015} is shown in Table \ref{tab:sensors}. All numbers reported are from actual sensor measurements, but we caution the readers that these comparisons are approximate and do not take into account process technology and other second order effects. Note that while the current ASP sensor is lower power, it is also much smaller resolution than the Sony image sensor. However, we argue that regardless of the image sensor, the power savings of turning on the ADC to digitize only edges will always be advantageous for embedded vision.

%

Since edge data is significantly smaller to transmit, ASPs can also save on the bandwidth of image data sent off the image sensor, thus alleviating an I/O bottleneck from image sensor to CPU. Prior work has shown that ASPs obtain a \textbf{bandwidth reduction of 10:1 or 90\%} for images by only storing non-zero coefficients of edges and using run-length encoding~\cite{wang2012compression}. For a traditional image sensor, 1.2 Mbits is needed to digitize 150Kpixels (384 $\times$ 384) at 8 bit resolution while ASPs only require 120Kbits. We refer readers to~\cite{wang2012compression} for more details about these circuit designs and their energy and bandwidth efficiency for ASP imaging.

\begin{table}[htbp]
	\begin{center}
   \includegraphics[width=1\linewidth]{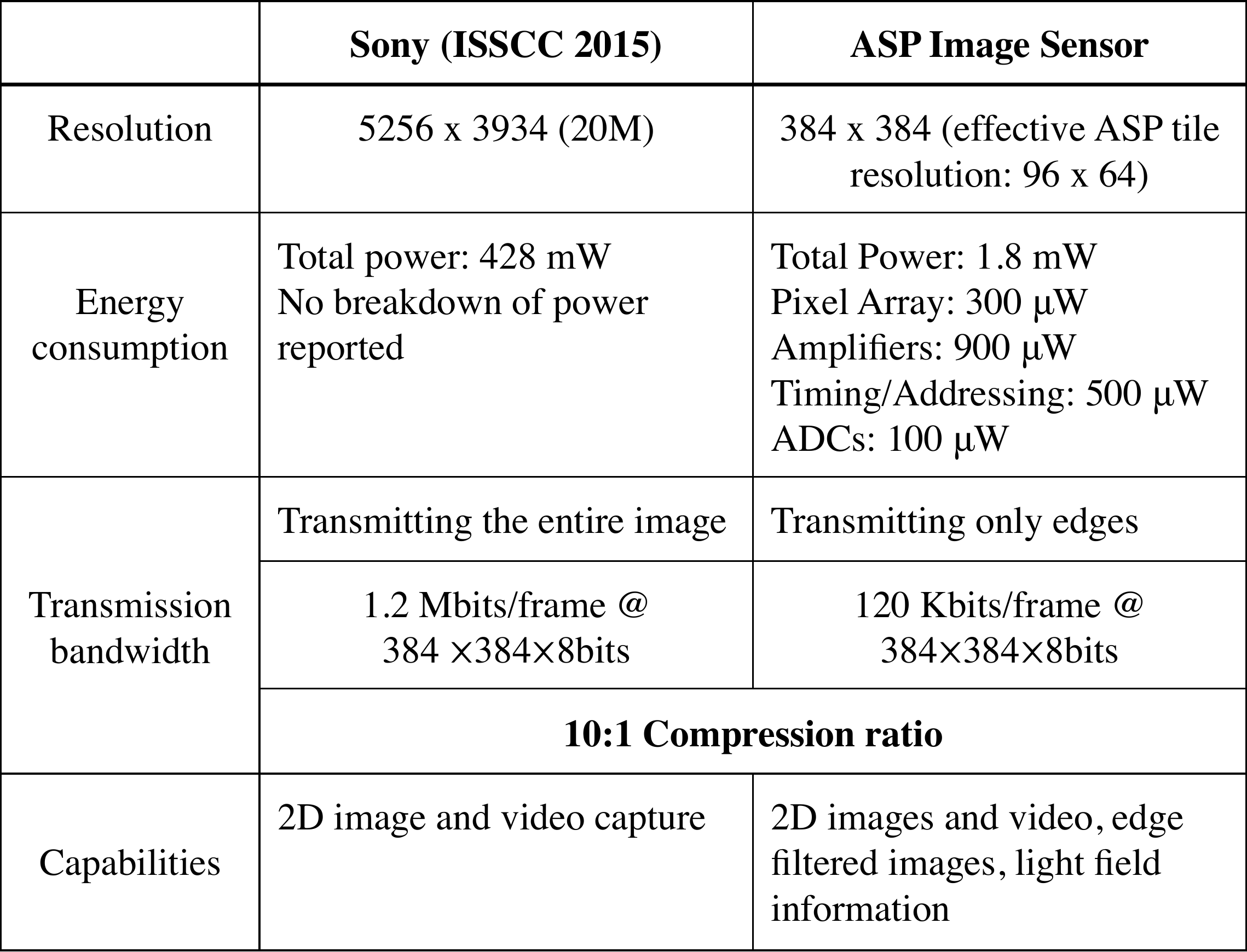}
\end{center}
   \caption{\textbf{Comparison of ASP image sensor~\cite{wang2012compression} and modern smartphone image sensor}~\cite{suzuki2015}.}
  \label{tab:sensors}%
\end{table}%

\subsubsection{Limitations of ASPs for Visual Recognition}
Some limitations with using ASPs for visual recognition include reduced image sensor resolution, low light efficiency, and depth-dependent edge filtering behavior. We outline these challenges and recent research to alleviate these issues. 

Since a tile of ASPs is required to obtain different edge filters, image sensor resolution is reduced by the tile resolution. It is not clear how small ASP pixels can be fabricated, especially since a few periods of diffraction gratings are needed for adequate signal-to-noise ratio (SNR) and to avoid edge effects. However, recent research in interleaved photodiode design has increased pixel density by 2$\times$~\cite{sivaramakrishnan2016design, sivaramakrishnan2011enhanced}. Reduced resolution may have an adverse effect on vision tasks~\cite{dai2015useful}, although no critical minimum resolution/spatial frequency threshold has been suggested for image sensors to capture.

ASP pixels can suffer loss of light through the diffraction gratings as low as 10\% relative quantum efficiency, which yields decreased SNR for differential edge responses. This in part explains the noisy visual artifacts present in the hardware prototype, and the need for large amounts of light in the scene. However, recent work in phase gratings~\cite{sivaramakrishnan2016design, sivaramakrishnan2011enhanced} have increased light efficiency up to 50\% relative quantum efficiency.

Finally, the optical edge-filtering behavior of ASPs is depth-dependent since the optical responses only work away from the focal plane with a large aperture camera~\cite{wang2012light}. This depth-dependence limits the application of ASPs to wide aperture systems with shallow depth-of-field, but also enables the potential for depth and light field information to be utilized as scene priors (which we do not explore in the scope of this paper).


\Section{Analysis}
	To analyze our proposed design and its tradeoffs, we developed a simulation framework to model both ASP image formation and CNNs. We simulate ASP image capture, and then propogate the resulting ASP edge images through the rest of the CNN. Typically this output data has dimensions W $\times$ H $\times$ D where there are D ASP filtered images, each of size W $\times$ H. We use the same input image resolution for both ASPs and baselines since we already accounted for image resolution in our normalized energy savings in Table \ref{tab:sensors}.

\begin{figure*}
\begin{center}
\includegraphics[width=.95\linewidth]{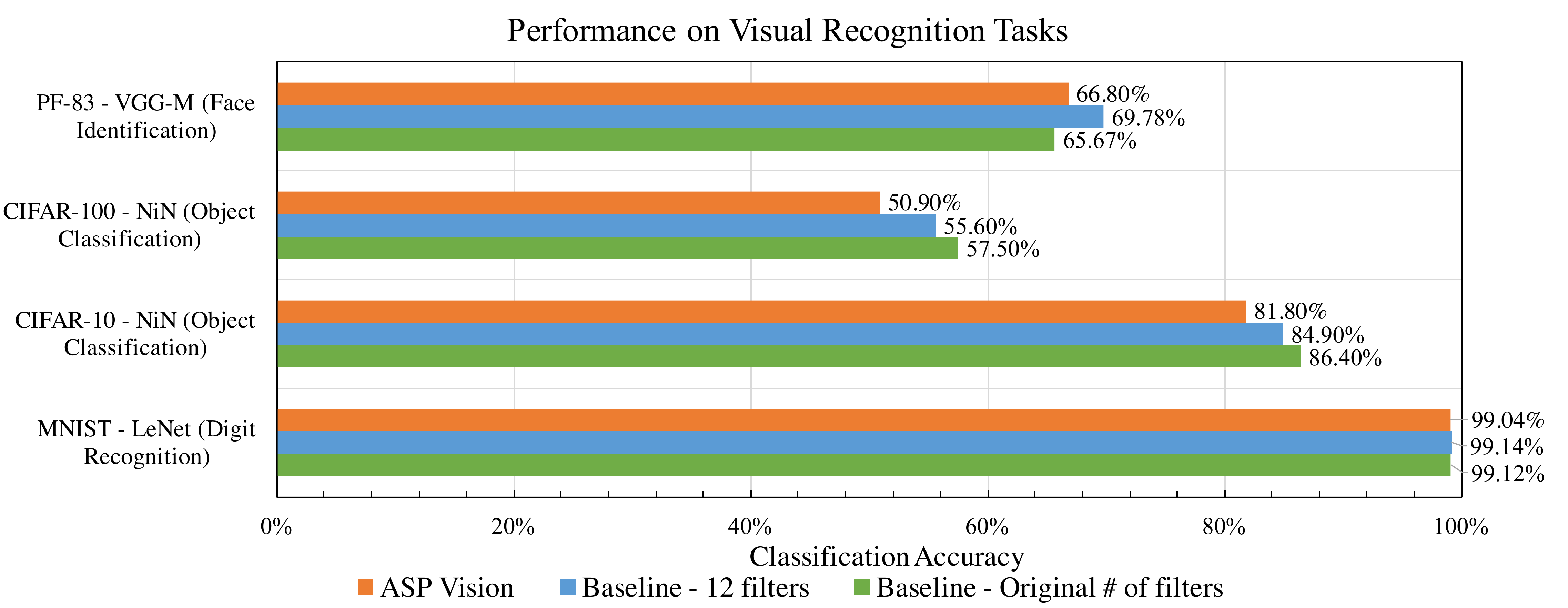} 
\end{center}
\caption{\textbf{ASP Vision Performance:} ASP Vision's performance on various visual recognition tasks, evaluated using three networks, LeNet~\cite{LeNet}, NiN~\cite{NiN} and VGG-M-128~\cite{VGG}, and over four different datasets: MNIST~\cite{lecun1998mnist},  CIFAR-10~\cite{krizhevsky2009learning}, CIFAR-100~\cite{krizhevsky2009learning}, and PF-83~\cite{becker2013evaluating}.}
\label{fig:performance}
\end{figure*}

For all our simulations, we use the ASP tile design of Figure \ref{fig:ring} which matches the existing hardware prototype of~\cite{wang2012compression}.  We use 12 out of 24 of the ASP filters with cosine responses ($\alpha = 0$) and low, medium, and high angular frequencies. The other 12 filters have sine responses ($\alpha = \pi/2$) which did not yield suitably different convolution outputs, and thus these matching input channels caused gradient exploding and convergence issues. Finally, since our prototype ASP system does not have color pixels, we report all baselines with respect to grayscale for performance. All our dataset results are summarized in Figure \ref{fig:performance} and discussed in the following subsection.

 We use MatConvNet~\cite{vedaldi15matconvnet} to perform deep learning experiments and train on a NVIDIA GeForce GTX TITAN Black GPU.

\begin{table*}
\begin{center}
\includegraphics[width=.8\textwidth]{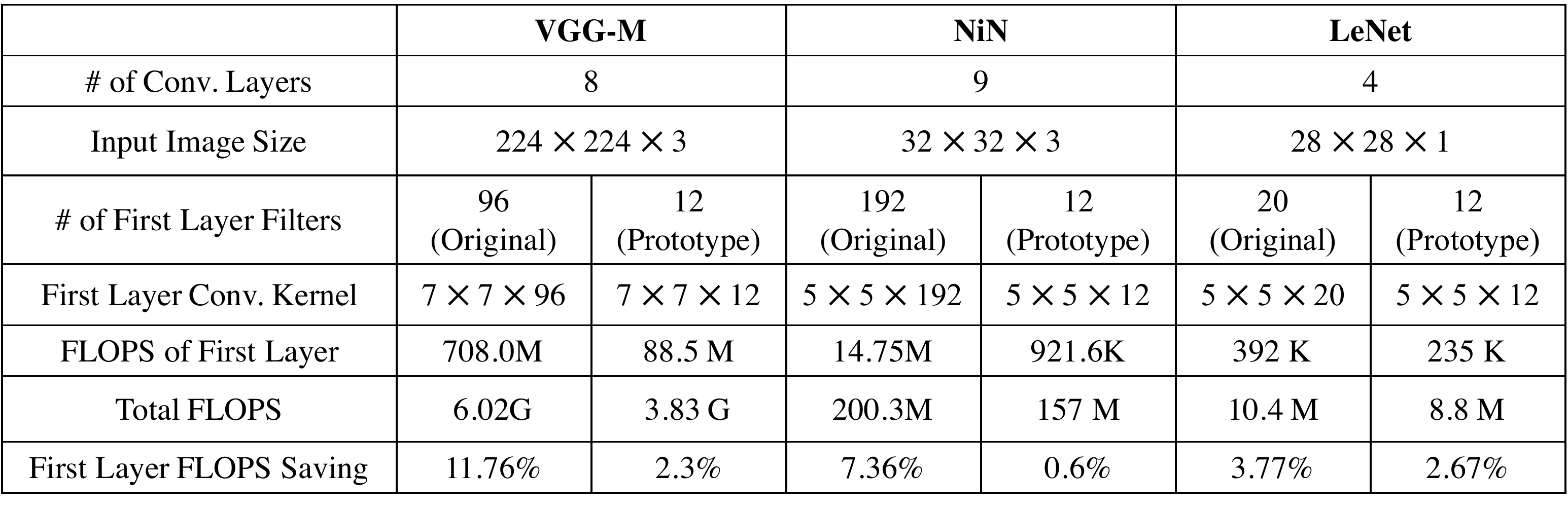} 
\end{center}
   \caption{\textbf{Network Structure and FLOPS:} Common CNN architectures such as VGG-M-128~\cite{VGG}, NiN~\cite{NiN}, LeNet~\cite{LeNet} are compared for the FLOPS savings from optically computing the first layer of these networks. The actual FLOPS savings for the working prototype ASP Vision system are also included.}
\label{tab:flops_calc}
\end{table*}

\subsection{Performance on Visual Recognition Tasks}
We first analyze the performance of ASP Vision across several visual recognition tasks to show the broad applicability of this system. The datasets we benchmark include MNIST~\cite{lecun1998mnist} for digit recognition, CIFAR-10/100~\cite{krizhevsky2009learning} for object recognition, and PF-83~\cite{becker2013evaluating} for face identification. 

For all experiments, we benchmark baselines with their original first layer number of filters (D) and also with $D = 12$ for a more fair comparison with ASP Vision when we analyze FLOPS in the next subsection. 

\textbf{MNIST:} Our first simulation involved digit recognition on MNIST, 60,000 training and 10,000 test images of size 28 $\times$ 28. For a baseline, we use LeNet~\cite{LeNet} which is a five layer CNN with both 20 and 12 first-layer filters to achieve 99.12\%and 99.14\% percent respectively. Using LeNet, ASP Vision achieved 99.04\% performance.

\textbf{CIFAR-10/100:} Our second simulation involved the CIFAR-10/100 data sets~\cite{krizhevsky2009learning} for object recognition with 50,000 training and 10,000 test images of size 32 $\times$ 32 (the 10/100 corresponds to the number of classes). Our baseline algorithm for these datasets was the Network in Network (NiN) structure~\cite{NiN} that uses CNNs with fully connected networks acting as inner layers. The baseline used both 92 and 12 first-layer filters to achieve respectively 86.40\% and 84.90\% percent on CIFAR-10, and 57.50\% and 55.60\% on CIFAR-100. Note again that these percentages are for grayscale images. ASP Vision achieved 81.8\% and 50.9\% respectively on CIFAR-10/100.

\textbf{PF-83:} Our final simulation on PF-83~\cite{becker2013evaluating} is an example of fine-grained classification to show that ASP features are transferable even for a difficult task like face identification (not to be confused with face verification or detection). The data consists of 13,002 images with size 256 $\times$ 256 with 83 classes of faces. Our baseline VGG-M-128 algorithm~\cite{VGG} achieved 65.67\% and 69.78\% percent on this data set with 192 and 12 first-layer filters respectively. Using ASP Vision, we achieved 66.8\% percent on PF-83. 

Across all datasets, ASP Vision was within 0.1-5.6\% of the baseline accuracies. Note that this comparable-to-slight degradation in performance comes with the energy savings of image sensing and transmission bandwidth by using ASPs. 

\subsection{FLOPS savings}
The FLOPS saved by ASP Vision is dependent on both the network architecture and the size of the input images. 

We first look at different CNN architectures and their potential savings from optically computing the first layer shown in Table \ref{tab:flops_calc}. Additionally, since we simulate only our hardware prototype of 12 filters, we compare the FLOPS of our prototype ASP Vision system with those of modified CNNs with a 12-filter first layer. This comparison results in lower FLOPS savings, but yields higher visual recognition performance. Using an ASP with more numbers of filters would allow more FLOPS savings when compared to CNNs with the equivalent number of first-layer filters. 

Secondly, FLOPS are input image size dependent as larger input image sizes will yield proportionally more FLOPS savings for an ASP Vision system. Even for a relatively deep network, the first layer still contributes a considerable amount of FLOPS if the input image is large. For example, the FLOPS of the first layer of GoogLeNet~\cite{szegedy2014going} is about 2.5\% of the total FLOPS.



\subsection{Noise analysis}
\begin{figure}[t]
\begin{center}
   \includegraphics[width=1\linewidth]{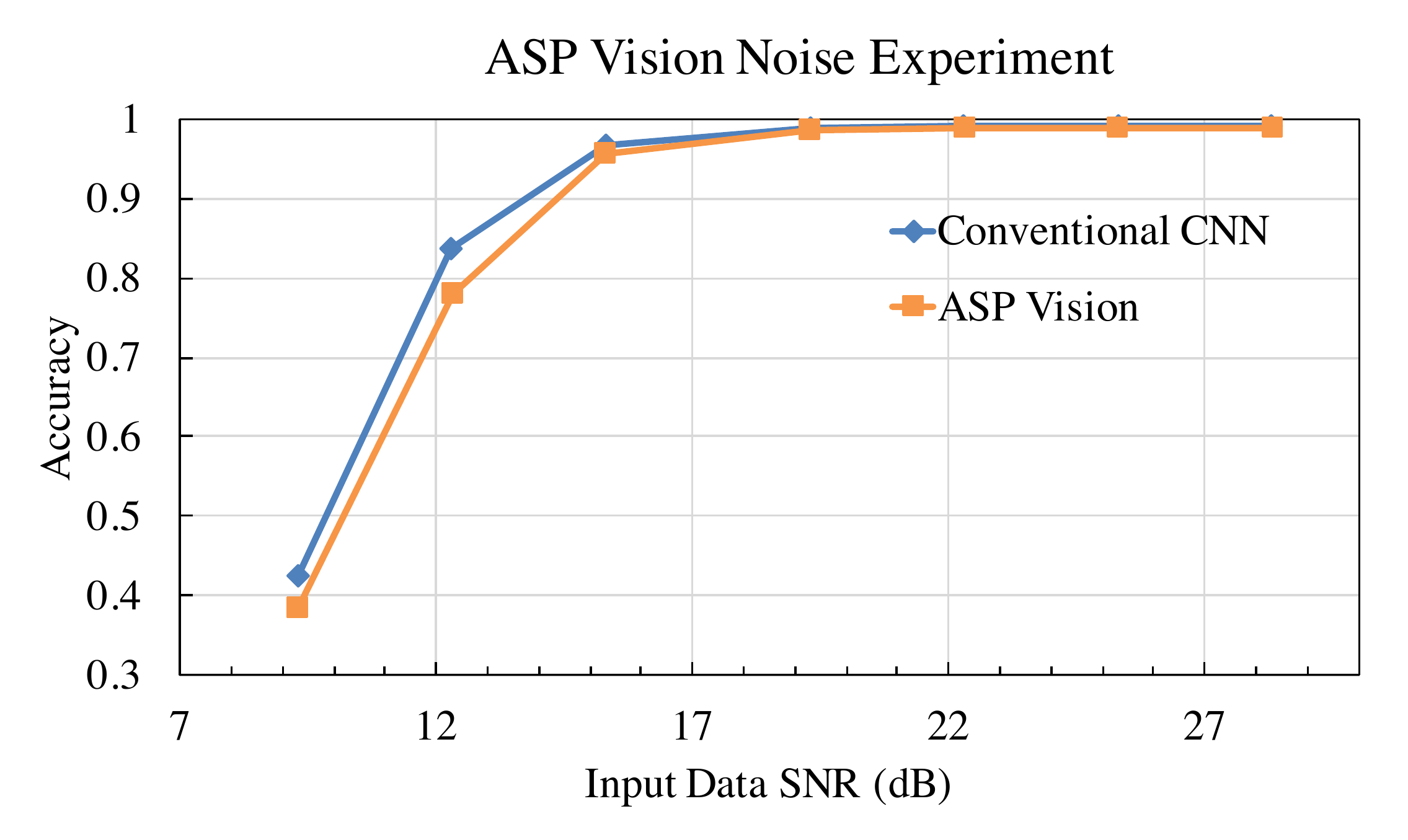}
\end{center}
   \caption{\textbf{ASP Vision Noise Analysis:} To explore the impact of noise to the performance of ASP Vision, we vary SNR from 9 dB to 28 dB and compared ASP Vision with baseline LeNet performance on MNIST.}
\label{fig:noisesweep}
\end{figure}

In Figure~\ref{fig:noisesweep}, we simulate the effects of additive white noise during image sensing for MNIST images. We compare ASP Vision versus the baseline LeNet with SNR varying from 9dB to 28dB. Note that at low SNRs, ASP Vision suffers more from accuracy degradation (9dB - 38.6\%, 12dB - 77.9\%) as compared to the baseline (9dB - 42.6\%, 12dB - 83.6\%). However, above 15dB SNR, both methods have high accuracy and are comparable. 

\subsection{ASP parameter design space}

We finally explore how choice of ASP parameters affects performance with the salient parameters being angular frequency $\beta$ and grating orientation $\chi$. We performed a coarse sweep of $\beta \in [5, 50]$, $\chi \in [-\frac{\pi}{2}, \frac{\pi}{2})$ for one filter on MNIST, and found no strong dependence on parameters and performance. 

We also ran sensitivity analysis on the parameter set by running 100 simulations using 6 randomized ASP filters each time on the MNIST dataset. We obtained a mean of 1.13\% error with a standard deviation of 0.13\%, which suggests there is no strong dependence of ASP parameters. This might be partly because the CNN learns to work with the filters it is given in the first layer.

\begin{figure}[t]
\begin{center}
   \includegraphics[width=\linewidth]{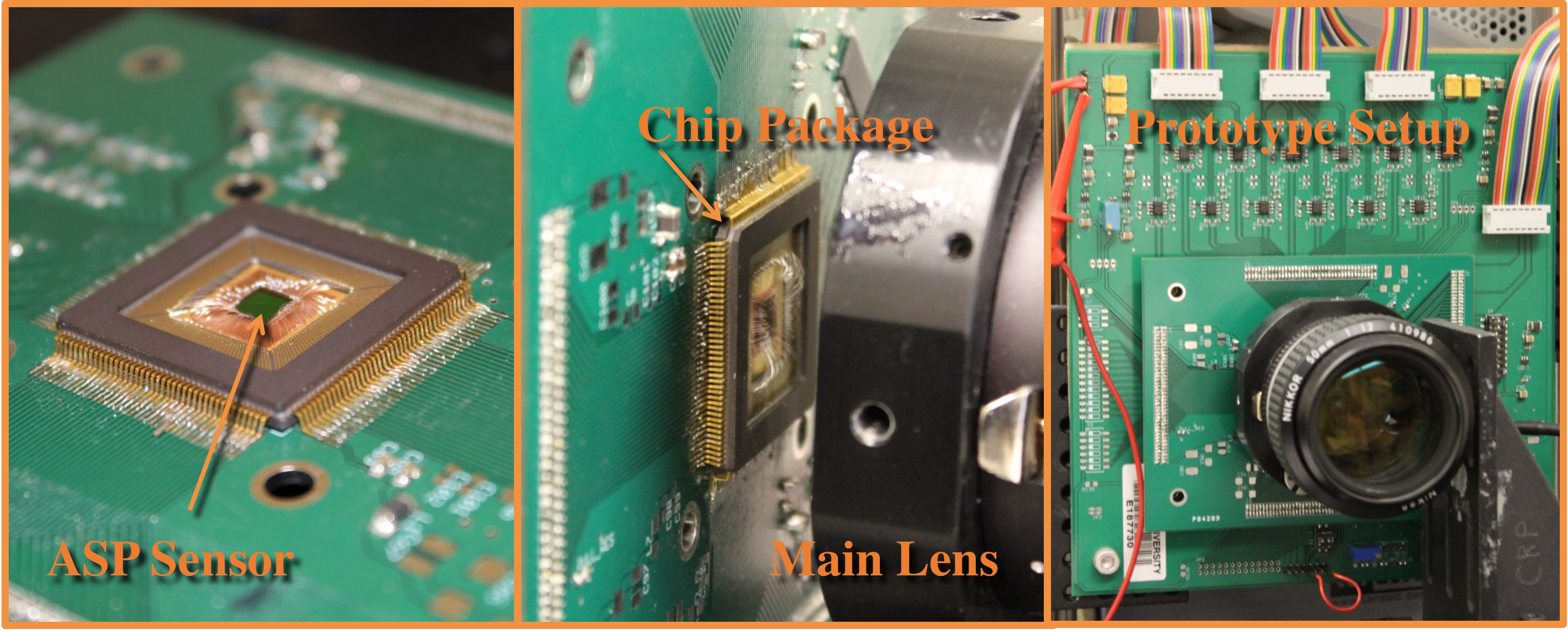}
\end{center}
   \caption{\textbf{ASP Camera Setup:} Working prototype with 5 mm x 5 mm CMOS ASP image sensor, F1.2 Nikon lens, and associated readout printed circuit board ~\cite{hirsch2014switchable, wang2012compression}.}
\label{fig:aspsetup}
\end{figure}

\Section{Hardware Prototype and Experiments}
	Finally, to completely validate our system design, we show results of classification on an existing camera prototype for digit and face recognition. We report mean validation accuracy and standard deviation for 20 trials with a random split of 85\% for training and 15\% validation. 

The prototype camera system is the same setup as used in~\cite{hirsch2014switchable, wang2012compression}.  A 5 mm $\times$ 5 mm CMOS image sensor was fabricated in a 180nm process, using a tile size of 4 $\times$ 6 ASPs with 10um pixels for a 64 $\times$ 96 resolution sensor. This sensor is placed behind a Nikon F1.2 lens for imaging small objects on an optical bench. See Figure~\ref{fig:aspsetup} for picture of our prototype camera.

In general, our prototype camera suffers from high noise even after a fixed pattern noise subtraction. This may be due to noise issues from the readout circuits or even from external amplifiers on the printed circuit board. This limits the aesthetics of the ASP edge images, but we still achieved high accuracy in visual recognition. Further circuit design such as correlated double sampling and fabrication in an industrial CMOS image sensor process could help alleviate these noise issues in the future. 

\begin{figure}[t]
\begin{center}
   \includegraphics[width=1.1\linewidth]{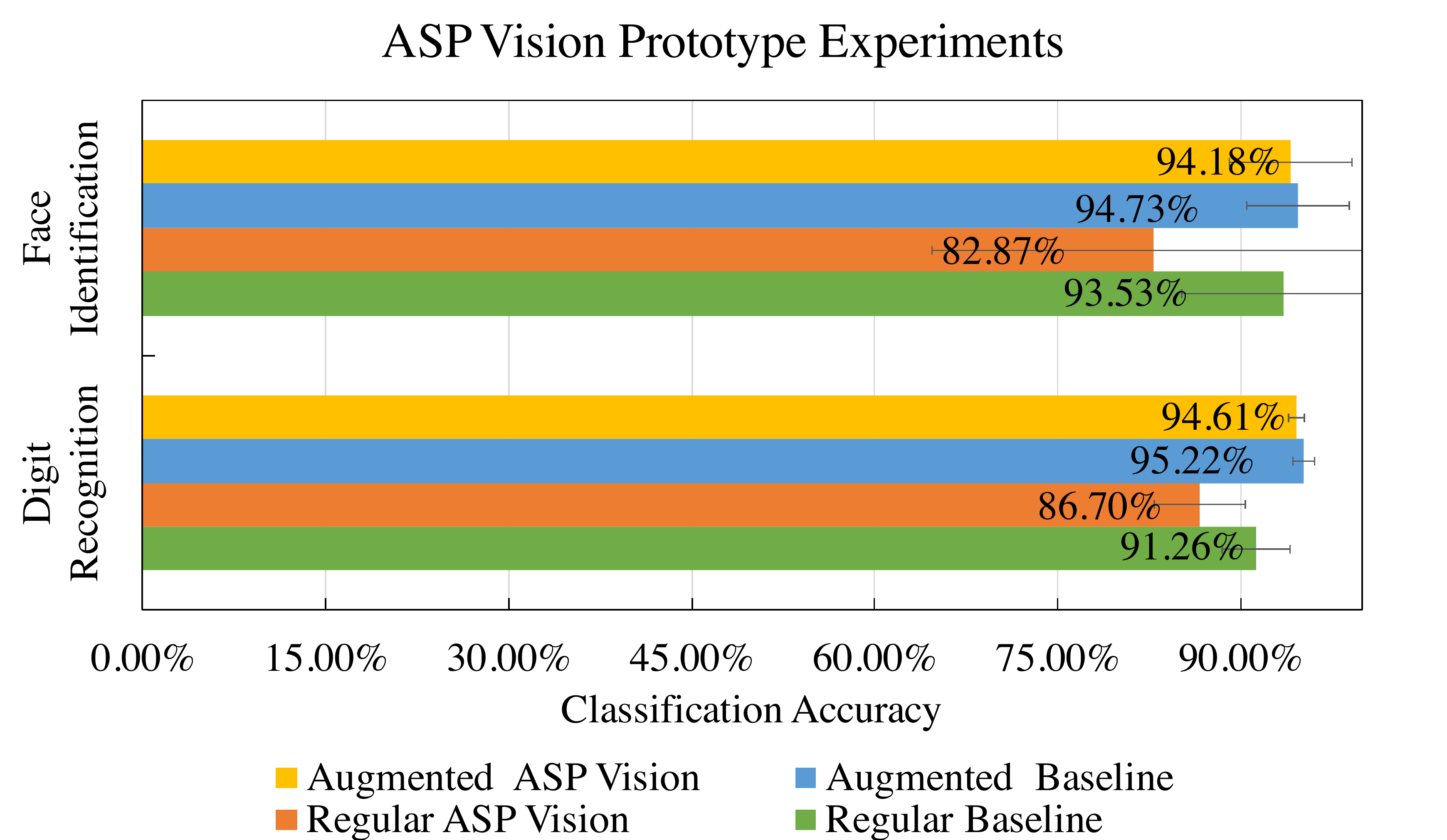}
\end{center}
   \caption{\textbf{ASP Vision Prototype Experiments:} Real-world digit recognition and face identification tasks were performed on ASP Vision prototype system. Accuracy and standard deviation for 20 trials are shown.}
\label{fig:realresults}
\end{figure}

\begin{figure}[t]
        \begin{center}
        \includegraphics[width=\linewidth]{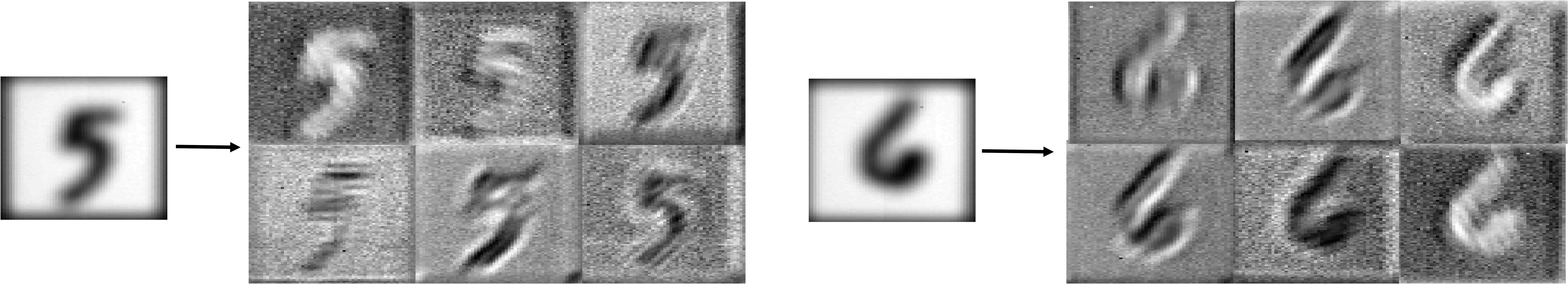}
        \end{center}
        \caption{\textbf{Digit Recognition:} Digits are captured by the ASP image sensor, 6 of 12 sample edge responses from the tile are shown. ASP Vision achieved \textgreater 90\% accuracy in digit recognition on this dataset.}
        \label{fig:digit}
    \end{figure}

\textbf{Digit Recognition:} Using a display with appropriate brightness approximately one meter away, we show images of the MNIST dataset, and capture ASP responses as shown in Figure~\ref{fig:digit}. We captured over 300 pictures of real digits to be used in our learning experiment. We also used linear shifts and rotations to augment the size of our dataset to 2946 images. For real data, the baseline LeNet algorithm performed 91.26\% with $\sigma = 2.77\%$ on the regular dataset, and 95.22\% with $\sigma = 0.87\%$ on the augmented dataset. ASP Vision achieved 86.7\% with $\sigma = 3.75\%$ on the regular dataset, and 94.61\% with $\sigma = 0.68\%$ on the augmented dataset.

\begin{figure}[t]
        \begin{center}
		\includegraphics[width=\linewidth]{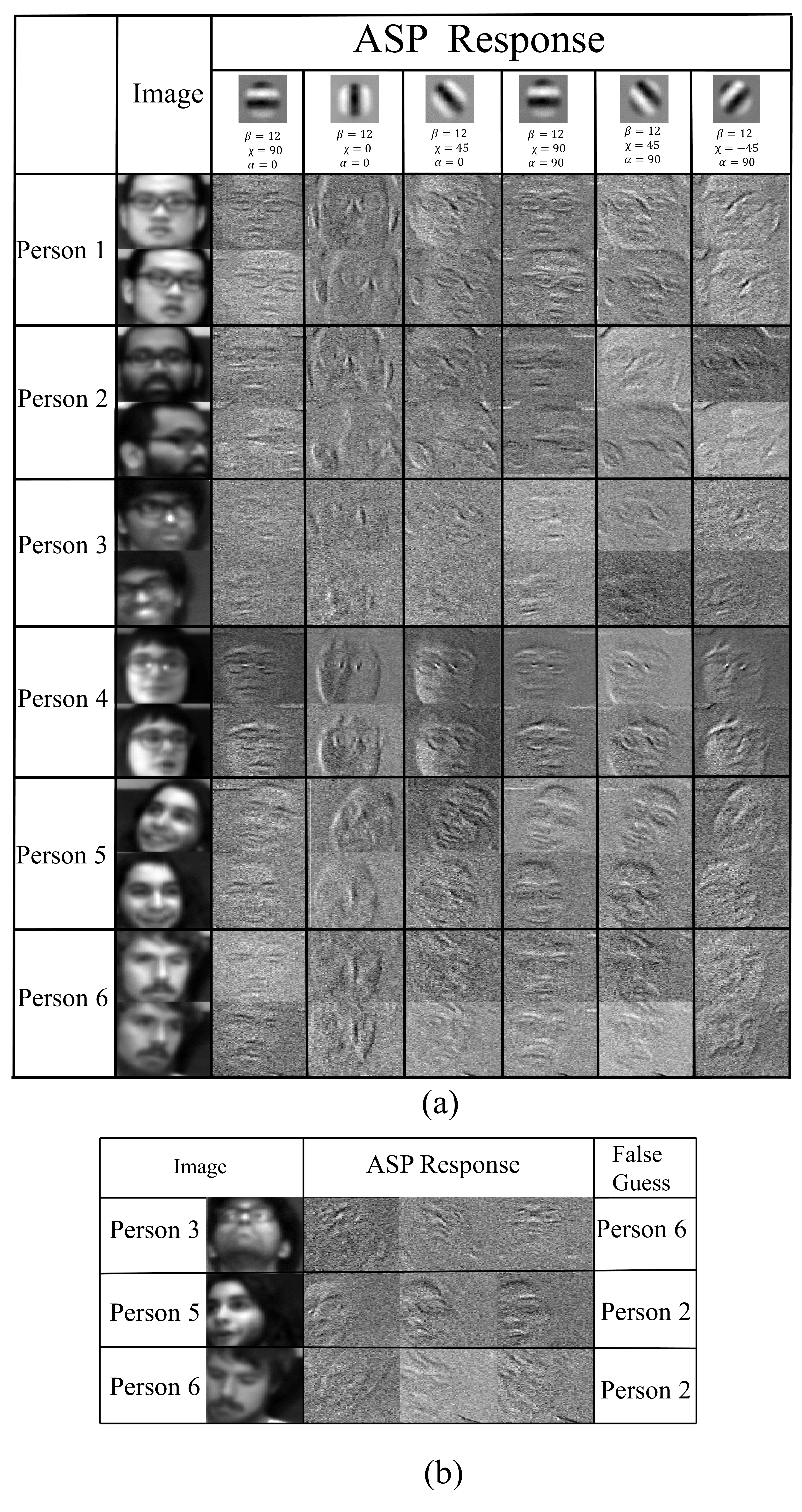} 
        \end{center}
		\caption{\textbf{Face Recognition:} 200 images of 6 subjects were captured in the lab. Edge responses for (a) correct and (b) misidentified identification is showed. ASP Vision achieved \textgreater 90\% accuracy for face identification.}        
		\label{fig:face}
    \end{figure}

\textbf{Face Identification:} To test face identification, we took 200 pictures of 6 subjects approximately 2.5 meters away in the lab, and the edge responses and example results and errors are visualized in Figure~\ref{fig:face}. We used dataset augmentation again to increase the dataset to 7200 pictures. For the baseline NiN, we achieved 93.53\% with $\sigma = 8.37\%$ on the regular dataset, and 94.73\% with $\sigma = 4.2\%$ on the augmented dataset. ASP Vision achieved 82.87\% with $\sigma = 18.12\%$ on the regular dataset, and 94.18\% with $\sigma = 5.04\%$ on the augmented dataset. 

ASP Vision performs about 5-10\% worse than baseline with regular data. After introducing linear shifts and rotations to augment the data, ASP Vision performs on par with conventional CNNs. These datasets may not generalizable and may exhibit underlying trends/bias due to the custom data acquisition. However, these results clearly show the feasibility of ASP Vision on a real working camera prototype.



\Section{Discussion}
Optically computing the first layers of CNNs is a technique that is not solely limited to ASPs. Sensors such as the DVS can compute edge features at low power/bandwidth~\cite{lichtsteiner2008128}, or using cameras with more general optical computation~\cite{zomet2006lensless} could capture convolutional features. In addition, it is not possible to hardcode additional convolutional layers optically in ASPs beyond the first layer, limiting the potential energy savings. Fully optical systems for artificial neural networks using holography~\cite{farhat1985optical, MIToptical, psaltis1990holography} or light waves in fiber~\cite{hermans2015towards} may achieve better energy savings.

We have presented an energy-efficient imaging system that uses custom Angle Sensitive Pixels for deep learning. We leverage energy savings and bandwidth reduction in ASPs while achieving good visual recognition performance on synthetic and real hardware data sets. We hope our work inspires sensor+ deep learning co-design for embedded vision tasks in the future.

\vspace{5mm}
\noindent \textbf{Acknowledgements:} We would like to thank Albert Wang for help designing the existing ASP prototype, and Dalu Yang and Dr. Arvind Rao for lending us the GPU. We also especially want to thank Dr. Eric Fossum and Dr. Vivienne Sze for pointing out corrections with regard to our energy comparison. We also received valuable feedback from Cornell's Graphics \& Vision group, Prof. David Field, Achuta Kadambi, Robert LiKamWa, Tan Nguyen, Dr. Ankit Patel, Kuan-chuan Peng, and Hang Zhao. This work was supported by NSF CCF-1527501, THECB-NHARP 13308, and NSF CAREER-1150329. S.J. was supported by a NSF Graduate Research Fellowship and a Qualcomm Innovation Fellowship. H.G.C was partially supported by a Texas Instruments Graduate Fellowship.


{\small
\bibliographystyle{ieee}
\bibliography{ASP-Vision}
}

\end{document}